\documentclass{article}
\usepackage{ijcai22}

\usepackage{times}
\usepackage{soul}
\usepackage{url}
\usepackage[hidelinks]{hyperref}
\usepackage[utf8]{inputenc}
\usepackage[small]{caption}
\usepackage{graphicx}
\usepackage{amsmath}
\usepackage{amssymb}
\usepackage{booktabs}
\usepackage{algorithm}
\usepackage{algorithmic}
\usepackage{color,xcolor}
\usepackage{multirow}
\usepackage{float}
\usepackage{caption}
\usepackage{newfloat}
\usepackage{listings}
\newcommand{\model}{{Regional Mask Guided Network}}
\newcommand{\modelfull}{{Regional Mask Guided Network (RMGN)}}
\newcommand{\modelshort}{{RMGN}}
\newcommand{\gate}{{regional mask}}
\newcommand{\gateshort}{{RM}}

\title{RMGN: A Regional Mask Guided Network for Parser-free Virtual Try-on}

\author{
Chao Lin$^1$\footnotemark[1]
\and
Zhao Li$^{2,3}$\footnotemark[1]
\and
Sheng Zhou$^2$\footnotemark[2]
\and 
Shichang Hu$^1$
\and
Jialun Zhang$^1$
\and
Linhao Luo$^4$
\and
\\
Jiarun Zhang$^5$
\and
Longtao Huang$^1$
\and
Yuan He$^1$
\affiliations
$^1$Alibaba Group, $^2$Zhejiang University, $^3$Link2Do Technology Ltd, \\ 
$^4$Monash University, $^5$University of California San Diego
\emails
\{linchao.lin, shichang.hsc, jay.zjl, kaiyang.hlt, heyuan.hy\}@alibaba-inc.com, \\
\{zhao\_li, zhousheng\_zju\}@zju.edu.cn, linhao.luo@monash.edu, jiz727@ucsd.edu
}

\graphicspath{{figures/}}
\begin{document}

\maketitle
\renewcommand{\thefootnote}{\fnsymbol{footnote}} 
\footnotetext[1]{Equal contribution.} 
\footnotetext[2]{Corresponding author.} 
\renewcommand{\thefootnote}{\arabic{footnote}}
\begin{abstract}

Virtual try-on (VTON) aims at fitting target clothes to reference person images, which is widely adopted in e-commerce. 
Existing VTON approaches can be narrowly categorized into Parser-Based (PB) and Parser-Free (PF) by whether relying on the parser information to mask the persons' clothes and synthesize try-on images. 
Although abandoning parser information has improved the applicability of PF methods, the ability of detail synthesizing has also been sacrificed.
As a result, the distraction from original cloth may persist in synthesized images, especially in complicated postures and high resolution applications.
To address the aforementioned issue, we propose a novel PF method named \modelfull. 
More specifically, a \gate\ is proposed to explicitly fuse the features of target clothes and reference persons so that the persisted distraction can be eliminated.
A posture awareness loss and a multi-level feature extractor are further proposed to handle the complicated postures and synthesize high resolution images.
Extensive experiments demonstrate that our proposed \modelshort\ outperforms both state-of-the-art PB and PF methods. Ablation studies further verify the effectiveness of modules in \modelshort\footnote{\url{https://github.com/jokerlc/RMGN-VITON}}. 
\end{abstract}

\begin{figure}[tbp]
    \centering
    \includegraphics[trim=0 0 0 0,clip,width=0.85\columnwidth]{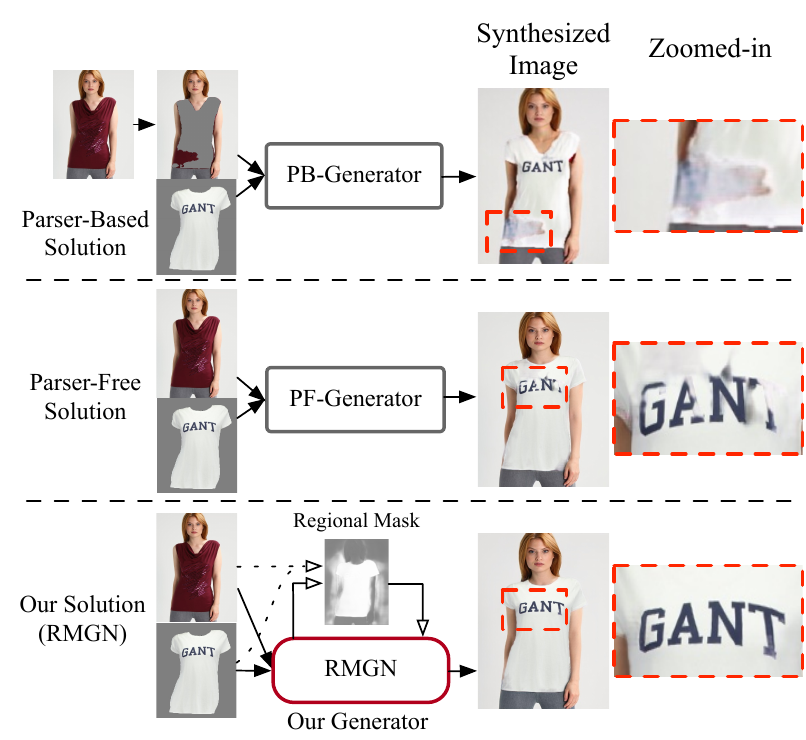} 
    \caption{The comparison of parser-based solution, parser-free solution, and our solution (\modelshort). The \modelshort\ can adaptively generate the \gate\ to integrate the advantage of both parser-based and parser-free solutions.
    } 
    \label{fig:intro}
\end{figure}

\section{Introduction}
\label{sec:intro}

With the flourish of e-commerce, fashion consumers demand a more realistic try-on experience during online shopping rather than viewing model images provided by merchants.
Such demand has recently been realized by virtual try-on (VTON) techniques where the target clothes (e.g. in-store clothes) are fitted onto the reference person images (e.g. consumers' images wearing arbitrary clothes). 

Early works on VTON have primarily focused on the Parser-Based (PB) solutions \cite{han2018viton,minar2020cp} consisting of a warp module and a generator module.
The PB first adopts the human parser \cite{gong2017look} to mask the reference person's cloth.
Then, the warp module takes a target cloth as input and deforms it to the reference person's body shape. The generator module synthesizes the try-on image by combining the cropped body parts with the target cloth.
Although PB methods have achieved great success recently, the dependence on accurate parser information has largely limited PBs' applicability in real-world circumstances \cite{issenhuth2020not}, especially when human postures are complex \cite{naha2020pose}. As shown in the first row of Figure \ref{fig:intro}, the over-reliance on human parsing makes PB solutions sensitive to inaccurate parsing results and prone to generate unrealistic images with obvious artifacts.

To alleviate this problem, Parser-free (PF) methods \cite{issenhuth2020not,ge2021parser} have drawn increasing attention recently by removing the need for parser information.
In WU-TON \cite{issenhuth2020not}, a student-teacher structure is proposed to guide the student model to mimic the ability of parser-based methods but without human parser. PF-AFN \cite{ge2021parser} redesigns the warp module to distill the appearance flows between the reference persons and target cloth images to achieve a more stable try-on effect.

However, most existing PF methods have mainly focused on learning a better parser-free warp module, while the impact on the generator is largely ignored.
More specifically, the generators in existing PF methods simply fuse the features of two inputs by concatenation. The feature distraction from the reference person is not considered and is persisted in the try-on result (as illustrated in Figure \ref{fig:intro}).
In fact, the generator module is critical for the existing PF framework. The generator not only synthesizes final try-on images that are the only basis for validating the performance of PF methods, but also provides guidance to the warp module in the cyclic training.

Although important, designing a generator under the parser-free setting is quite challenging. 
First, the style of the target cloth may be different from the reference person's cloth, how to eliminate the distraction from the reference person's cloth in synthesized images without parsing guidance?
Second, in practice, the body posture may be complicated, how to adjust the target cloth to fit complicated human postures? 
Last but not least, fashion consumers demand high resolution try-on images, how to generate high-quality try-on results in high resolution to meet their requirements?

To tackle the aforementioned challenges, this work proposes a novel parser-free VTON method called \modelfull. 
The \modelshort\ model can generate real-looking pictures in high resolution, eliminate the distraction of the the reference persons' cloth, and handle complex patterns and postures. Specifically, (1) a \textit{multi-level feature extractor} is proposed to separately extract features from reference persons and target clothes, 
which ensures the \modelshort\ to generate high resolution images with more details.
(2) a novel \textit{\gate} is proposed to explicitly fuse the features of the target cloth and the reference person without parser information. Therefore, both the advantages of PBs and PFs can be incorporated in a unified framework.As illustrated in the third row of Figure \ref{fig:intro}, this helps \modelshort\  better retain the the color and pattern of the target cloth.
(3) a \textit{posture awareness loss} is also proposed to focus on the structure and margin. This enables \modelshort\ to achieve more stable results with respect to a variety of human postures and body types.

Noticeably, to the best of our knowledge, we are the first who proposed mask guided parser-free model in VTON. It is different from previous mask-guided methods which usually relied on the inputted segmentation. For example, GAN-based method \cite{gu2019mask} used inputted mask to change its semantic context to customize portrait details. While \modelshort\ can extract features of clothes and person to adaptively generate the \gate\ without any human effort. In summary, our contributions are listed as follows:
\begin{itemize}
    \item We propose a novel parser-free model called \modelfull\ incorporating the advantages of both PBs and PFs, which is able to generate high-quality try-on images in high resolution.
    \item We introduce a posture awareness loss function to allow our proposed model to deal with complex postures that most of the existing models failed to handle.
    \item Extensive experiments on two public datasets demonstrate that \modelshort\ outperforms other baselines and achieves the state-of-the-art performance. 
\end{itemize}

\section{Related Work}
\label{sec:related}
Existing 2D VTON methods, due to their efficiency, have drawn increasing attentions, which can be further divided into two: parser-based and parser-free solutions. 

\noindent\textbf{Parser-based VTON.} VITON \cite{han2018viton} proposed a coarse-to-fine framework that transfers a in-shop cloth to the corresponding region of a reference person. CP-VTON \cite{wang2018toward} adopted a TPS transformation to obtain a more robust and powerful alignment. But VITON and CP-VTON only focus on the cloth region. ACGPN \cite{yang2020towards} is the first in the field which predicts the semantic layout of the reference person to determine whether to generate or preserve its image content. To generate better results in high resolution, VITON-HD \cite{choi2021viton} proposed an ALIAS normalization as well as a generator to handle the misaligned areas and preserve the details. Consequently, it can synthesize 1024x768 outputs. However, all the above solutions require accurate human parsing while parsing errors will lead to try-on images with obvious artifacts.

\noindent\textbf{Parser-free VTON.} Recently, WUTON \cite{issenhuth2020not} put forward a parser-free virtual try-on solution using a student-teacher paradigm but bounds the image quality of the student to the parser-based model. To address this problem, PF-AFN \cite{ge2021parser} proposed a ``teacher-tutor-student'' knowledge distillation scheme, 
which distills the appearance flow between person and cloth images for high-quality generation. But, these two parser-free solutions ignore the fact that clothes on the reference person will disturb the generator module leading to sub-optimal try-on results, especially in high resolution. TryOnGan \cite{lewis2021tryongan} focused on changing clothes between persons, while our RMGN aims to fit a given cloth onto the person image.

\noindent\textbf{Mask Guided Image Synthesis.} Masks guided methods have been proposed to address many tasks in computer version \cite{zhu2020sean,lee2020maskgan}. FaceShifter \cite{li2019faceshifter} proposed an adaptive attention generator to adjust the effective region of the identity and attribute embedding with an attention mask. SEAN \cite{zhu2020sean} proposed semantic region-adaptive normalization for Generative
Adversarial Networks conditioned on segmentation masks. LGGAN \cite{tang2020local} adopted masks as guidance for local scene generation. However, the \gate\ is automatically generated in \modelshort\ without any human effort.

\section{Preliminary}
\label{sec:pre}
The commonly used parser-free framework \cite{ge2021parser} contains a pre-train stage and a parser-free training stage. The first stage pre-trains a PB-Warp module and a PB-Generator module, while the second stage distills a PF-warp module from the PB-warp and optimize a PF-generator.

The notations used in this paper are introduced as follows. The $P$ denotes the reference person image which we intend to change a given cloth on, $P_c$ is the person image with masked body region, and $\hat{P}$ is the synthesized image. $I$ and $I_t$ denote arbitrary clothes that are used to create fake triplets and the target cloth we want to fit onto the reference person. 
$I'_*$ denotes the deformed clothes warped to the person's body shape, and $\overline{I'_t}$ is the deformed cloth cropped from $P$ and severed as the ground truth for the warp modules. 
The fake image $\widetilde{P}$ is a synthesized image generated from $P_c$ and $I'$. The ($\widetilde{P},I_t,P$) is the fake triplet used for PF module training.

Knowledge distillation \cite{ge2021parser} is a widely used method for the PF-Warp module. By minimizing the distillation loss $\mathcal{L}_d$, the PF-Warp module could mimic the warping ability from PB-Warp but without using any parser information. 
The details of $\mathcal{L}_d$ can be found at Appendix.

\noindent\textbf{Problem Definition.} Given a reference person $P$, and a target cloth $I_t$, our goal is to fit the target cloth onto the reference person to synthesize the try-on image $\hat{P}$.

\begin{figure*}[]
    \centering
    \includegraphics[trim=1cm 3cm 3.3cm 2cm,clip,width=0.95\textwidth]{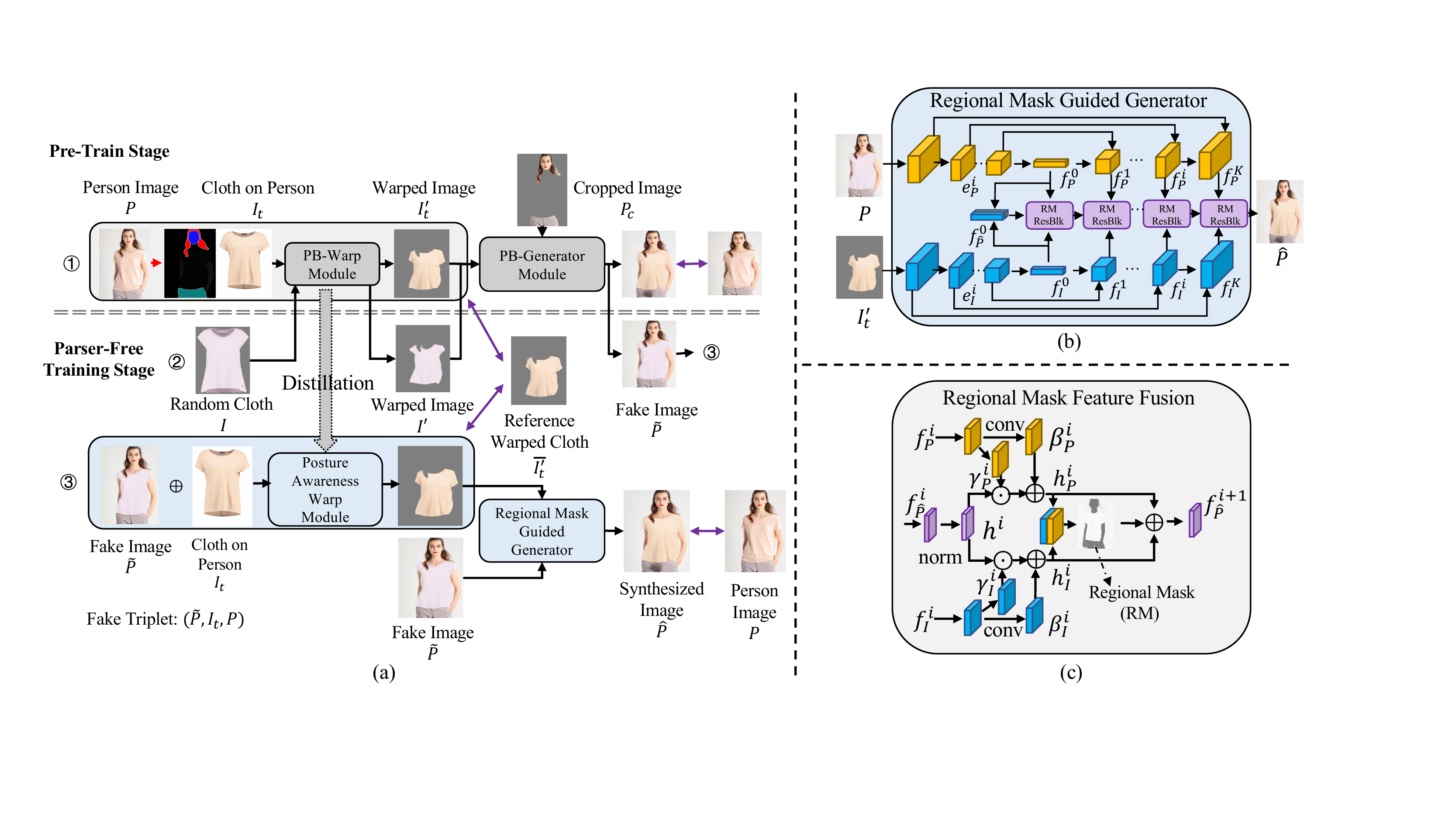} 
    \caption{(a) The overall framework of the proposed \modelfull. (b) the detail of the novel Regional Mask Guided Generator. (c) the detail of the Regional Mask Feature Fusion.
    } 
    \label{fig:AMGN}
\end{figure*}

\section{\model}
\label{sec:approach}

In this paper, we focus on the parser-free training stage, while improving the warp module and proposing a novel generator. The whole framework of \modelfull\ is illustrated in the left part of Figure \ref{fig:AMGN}.


\subsection{Posture Awareness Warp Module}
\label{sec:warp}

The warp module aims to deform the clothes to fit the human pose and body shape while preserving the clothes' details. 
Following AFWM \cite{ge2021parser}, we adopt different convolution layers to scale the original image to different sizes, on which the cloth is wrapped to generate the final warped cloth.
This module is optimized under the guidance of a pixels-wise loss (first-order constrain) $\mathcal{L}_f$, and a smooth loss (second-order constrain) $\mathcal{L}_{sec}$, which are written as
\begin{equation}
    \setlength\abovedisplayskip{1pt}
    \setlength\belowdisplayskip{1pt}
    \begin{gathered}
        \mathcal{L}_f = ||I'_t - \overline{I'_t} ||\\
        \mathcal{L}_{sec} = \sum_{i=1}^L \sum_j\sum_{\pi\in \mathcal{N}_j} \mathcal{P} (f^i_{j-\pi} + f^i_{j+\pi}- 2f^i_{j}),
    \end{gathered}
\end{equation}
where $f^i_j$ denotes the feature value of $j$-th point on the $i$-th scale feature map, $\mathcal{N}_j$ denotes the set of horizontal, vertical, and both diagonal neighborhoods around the $j$-th point; $\mathcal{P}$ denotes the generalized Charbonnier loss \cite{sun2014quantitative}.

Although the warped clothes are aligned with the ground truth, the human posture is not paid enough attention, which may deteriorate the warped results, especially when the posture is complicated.
To this end, a \textit{posture awareness loss} $L_W$ is proposed to assure the PF-Warp module deforms the clothes with more emphasis on postures.
For each target cloth $I_t$, we randomly generate a set of fake images $\widetilde{\mathbb{P}}$ with different sampled clothes but in the same posture. 
Then, using the fake image set and the target cloth, we adopt PF-Warp to generate a set of warped clothes $\mathbb{I}'_t$ for $\mathcal{L}_W$ optimization, which can be formulated as
\begin{equation}
    \setlength\abovedisplayskip{1pt}
    \setlength\belowdisplayskip{1pt}
    \begin{gathered}
        I'_{t,j} = \text{PF-Warp}(\widetilde{P}_j, I_t),\ \widetilde{P}_j \in \widetilde{\mathbb{P}}\\
        \mathcal{L}_W = \frac{1}{|\mathbb{I}'_t|}\sum_{ I'_{t,j} \in \mathbb{I}'_t} \lambda_{f}\mathcal{L}_{f}^j + \lambda_{sec}\mathcal{L}_{sec}^j + \lambda_{d}\mathcal{L}_{d}^j,
    \end{gathered}
\end{equation}
where $\lambda_{f}$, $\lambda_{sec}$, $\lambda_{d}$ denote the weight of the corresponding loss, and $\mathcal{L}_{f}^j$, $\mathcal{L}_{sec}^j$, $\mathcal{L}_{d}^j$ denote the aforementioned loss calculated from $I'_{t,j}$ and $\overline{I'_t}$.
Through the posture awareness loss, the warp module can better deform the target cloth to persons' postures regardless of different wearing clothes, which is also verified in the experiment parts.

\subsection{Regional Mask Guided Generator}
\label{sec:gen}

The generator module takes warped images and reference person images as input to synthesize final try-on images. 
Prior PF-generator has failed to eliminate the distraction from reference persons' clothes, as a result, only low-quality try-on images can be generated in high resolution. 
To synthesize high resolution images with more details of target clothes and reference persons, we propose a two-way feature extractor and progressively fuse the extracted features into final try-on images through a one-way generation process. 
The details of the proposed Regional Mask Guided Generator are illustrated in the upper right of Figure \ref{fig:AMGN}.

\noindent\textbf{Multi-level Feature Extractor.}
In the high resolution VTON, the features at different levels should be considered since the shallow level features imply local information (e.g., color and texture) and the higher level features contain more global information (e.g., body and cloth region) \cite{zeiler2014visualizing}. 
To make full use of these information, we propose a $K$-layer encoder-decoder to respectively capture features at different levels, which can be formulated as
\begin{equation}
    \setlength\abovedisplayskip{1pt}
    \setlength\belowdisplayskip{1pt}
    \begin{gathered}
        x_*^{i+1} = \text{DeConv}(f_*^{i}) \\
        f_*^{i+1} = x_*^{i+1} + e_*^{K-i},
    \end{gathered}
\end{equation}
where $f_*^i$ denotes the $f_P^i$ or $f_I^i$ on $i$-th decoder layer when $i=0$, $f_*^0 = e_*^K$, and $e_*^{K-i}$ denotes features on ($K-i$)-th encoder layer. We adopt a short connection to retain multi-level features between the encoder and decoder.

Then, to refine the try-on results gradually, we propose a sub-module called \gateshort-ResBlk and use it to fuse the extracted features layer-by-layer. \gateshort-ResBlk is made up of several Regional Mask Feature Fusion blocks in the form of residual connection \cite{he2016deep}. The details of this sub-module will be illustrated in the Appendix.

\noindent\textbf{Regional Mask Feature Fusion.}
To alleviate the feature distraction faced by the prior PF-generator and generate photo-realistic images in high resolution, we introduce a Regional Mask Feature Fusion block guiding the feature fusion explicitly.
As illustrated in the bottom right of Figure \ref{fig:AMGN}, we adopt a de-nomalization mechanism \cite{park2019semantic} to preserve important features.
First, it passes the $f_{\hat{P}}^{i}$ through a normalization layer \cite{ulyanov2016instance} to get an activation feature map $h^i$. Then, it adopts two independent convolution layers with window size 1 as the kernel functions, which project the representations of persons and clothes into $\gamma_*^i$ and $\beta_*^i$. Finally, the activation feature map $h^i$ is interacted with them respectively, which can be formulated as
\begin{equation}
    \setlength\abovedisplayskip{1pt}
    \setlength\belowdisplayskip{1pt}
    \begin{aligned}
        h_P^i &= h^i \odot \gamma_P^i + \beta_P^i \\
        h_I^i &= h^i \odot \gamma_I^i + \beta_I^i,
    \end{aligned}
\end{equation}
where $\odot$ denotes the element-wise multiplication. As a result, $h^i$ can be de-normalized into $h^i_P$ and $h^i_I$ with important features activated and preserved for high resolution try-on images.

As we discussed in the introduction, the prior PF-Generator suffers from the distraction of the reference person's cloth. We find that the details of the synthesized images are either from the reference person or the target cloth, features of the reference person and target cloth at the same region are often exclusive during the feature fusion. Their features could interfere with each other deteriorating the quality of the synthesized image. Therefore, we propose a \textit{\gate\ (\gateshort)} to explicitly select incoming features and guide the feature fusion.

The $h^i_P$ and $h^i_I$ are concatenated together to pass a convolution layer and a sigmoid activation function to generate the final \gate\ $M^i$, which can be formulated as
\begin{equation}
    M^i = \sigma(\text{Conv}(h^i_P||h^i_I)),
\end{equation}
where $||$ denotes the concatenation operation, and $\sigma(\cdot)$ denotes the sigmoid function that normalizes the attention values between 0 and 1.

The \gate\ $M^i$ can explicitly select the incoming features without any supervision. 
Therefore, the Regional Mask Feature Fusion can be formulated as
\begin{equation}
    f^{i+1}_{\hat{p}} = (1-M^i)\odot h^i_P + M^i\odot h^i_I.
\end{equation}

It is worth noting that unlike the traditional human parsing used by PBs, we integrate \gate\ with the try-on generation process in an end-to-end manner. By doing which, \gate\ incorporates the advantages of both PBs and PFs. More precisely, \gate\ is trained to spontaneously focus more on the regions that are important to the try-on effect, such as new skin regions where used to be collars and sleeves. We will further visualize and evaluate the effect of \gate\ in the experiment section.

\subsection{\modelshort\ Optimization}
\label{sec:loss}

The whole optimization process of proposed \modelshort\ is conducted under the cyclic training framework. We first use the reference person images and arbitrary clothes to generate fake images. Then, we perform the parser-free optimization on the fake triplet ($\widetilde{P},I_t,P$), where the reference person images are used as the ground truth in return to optimize \modelshort. 

For the generator module, its loss function consists of a pixel-wise $\mathcal{L}_f$ loss and a perceptual loss \cite{johnson2016perceptual} $\mathcal{L}_p$ to encourage the visual similarity between the reference try-on images, which can be formulated as
\begin{equation}
    \setlength\abovedisplayskip{1pt}
    \setlength\belowdisplayskip{1pt}
    \begin{gathered}
        \mathcal{L}_f = ||\hat{P} - P || \\
        \mathcal{L}_p = \sum_m ||\phi_m(\hat{P}) - \phi_m(\hat{P}) ||, 
    \end{gathered}
\end{equation}
where $\phi_m(\cdot)$ indicates the feature map on $m$-th layer of VGG-19 which is pre-trained on ImageNet.

Similar to the PF-Warp module training, for each target cloth $I_t$, we adopt the fake images set $\widetilde{\mathbb{P}}$ and warped images set $\mathbb{I}'_t$ to optimize the Regional Mask Guided Generator. The final generator loss can be formulated as
\begin{equation}
    \setlength\abovedisplayskip{1pt}
    \setlength\belowdisplayskip{1pt}
    \begin{gathered}
        \hat{P_{j}} = \text{\gateshort-Generator}(I'_{t,j},\widetilde{P}_j) \\ 
        \mathcal{L}_G = \sum_{j} \lambda_{f} \mathcal{L}_f(\hat{P_{j}}, P) + \lambda_{p} \mathcal{L}_p (\hat{P_{j}}, P),
    \end{gathered}
\end{equation}
where $\lambda_{1}$ and $\lambda_{p}$ are the corresponding loss weights.

Finally, we can optimize the parameters $\theta$ of \modelshort\ by the following optimization function
\begin{equation}
    \setlength\abovedisplayskip{1pt}
    \setlength\belowdisplayskip{1pt}
    arg \mathop{Min}\limits_{\theta} \mathcal{O(\theta)} = \mathcal{L}_W + \mathcal{L}_G.
\end{equation}

The implementations are available in the supplementary files and will be released upon acceptance.

\begin{figure*}[]
    \centering
    \includegraphics[trim=0cm 0cm 0cm 0cm,clip,width=0.95\textwidth]{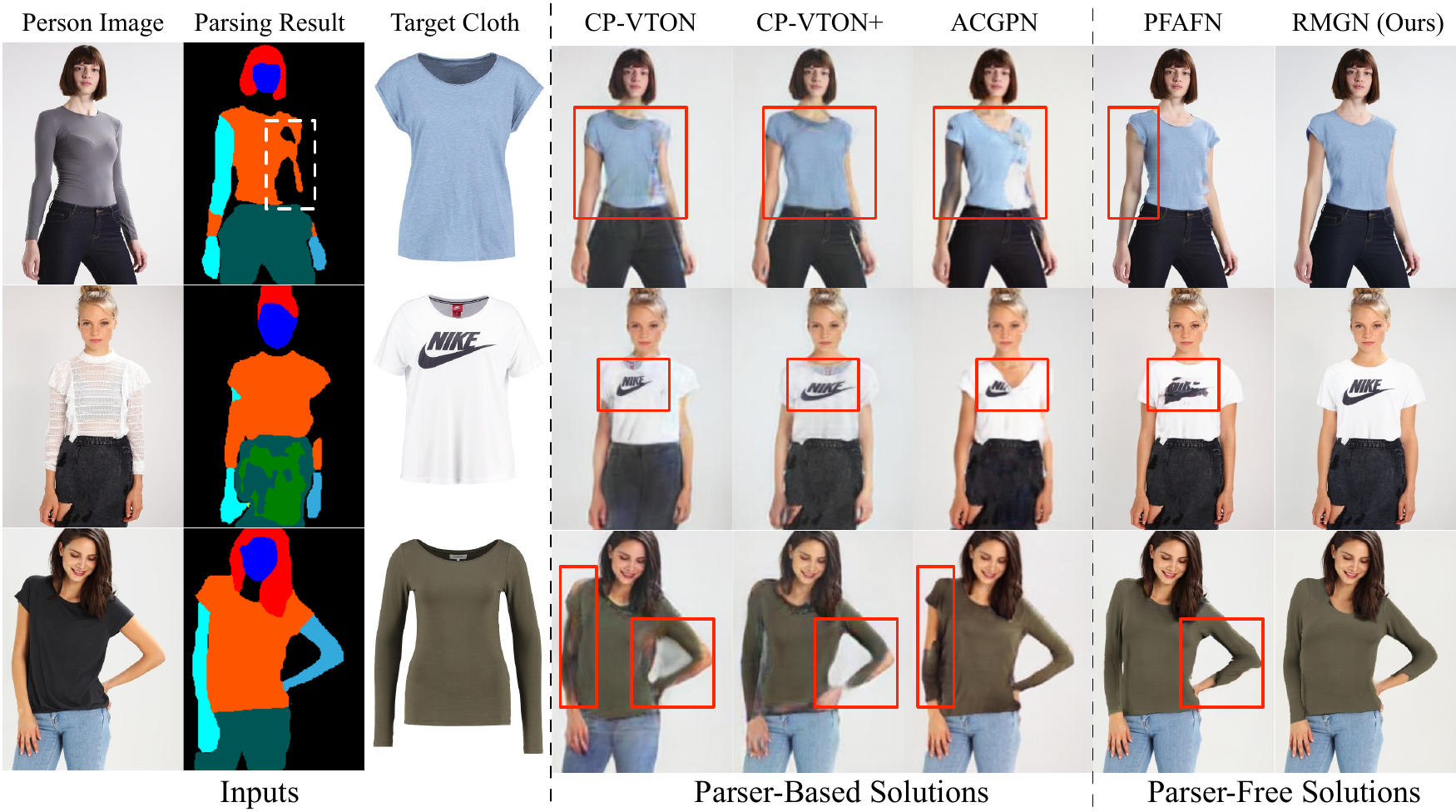} 
    \caption{Visual comparison on 512 $\times$ 384 VITON dataset where parsing results are only used by parser-based methods. Comparing with other methods, in high resolution, our method better handles the situations of inaccurate parsing results (row 1), distraction from original cloth (row 2), and complex human postures (row 3).}
    \label{fig:viton}
\end{figure*}

\section{Experiments}
\label{sec:experiment}

We conduct experiments on two public datasets: VITON \cite{han2018viton} and MPV \cite{dong2019towards}, which are widely applied by recent researches in this field \cite{wang2018toward,yang2020towards,ge2021parser}. VITON contains 19,000 front women images and their corresponding top cloth images with the resolution of both 256 $\times$ 192 and 512 $\times$ 384. Original MPV contains 35,687/13,524 person/cloth images at 256 $\times$ 192 resolution. Respectively, we filtered out 14,221/2032 training/testing pairs from VITON, and 11,032/2,390 training/testing pairs with clear clothes and person details from MPV to construct the MPV-Sub.


\subsection{Qualitative Results}
To evaluate the performance of proposed \modelshort, we compare it against both PB methods: CP-VTON \cite{wang2018toward}, CP-VTON \cite{minar2020cp}, ACGPN \cite{yang2020towards}, and the PF methods: WUTON \cite{issenhuth2020not}, PFAFN \cite{ge2021parser}, on two datasets with different resolutions. 
Figure \ref{fig:viton} illustrates the results on VITON dataset in high resolution ($512 \times 384$) and the results on MPV are illustrated in the Appendix. More specifically, we would like to show the performance of all methods in three challenging situations:

\noindent\textbf{Inaccurate Parsing Result.} 
When the segmentation information provided by human parser is inaccurate, previous PB methods would leave noticeable artifacts on the synthesized images. In the example showed in the first row of Figure \ref{fig:viton}, the parser failed to recognize the cloth areas labeled out by the white dash rectangle. 
As a result, there are large areas of residuals on the synthesized person's chest and waist region (the red rectangle areas). Though PFAFN's cloth area would not be affected by the inaccurate parsing, on the other hand, it fails to recover the skin details distracted by the original cloth without parsing guidance. Our \modelshort\ incorporates the advantages of both PB and PF, and generates images with both clear cloth and skin region.

\noindent\textbf{Distraction from Original Cloth.}
The detailed content of the target clothes, such as the clothes' logos and patterns, can be influenced by the original cloth. The second row of Figure \ref{fig:viton} is a representative example where the distraction is maximized as the original cloth is in different style from the target cloth. Comparing the red rectangle area, PFAFN performs the worst where the logo is mixed with other patterns coming from the original cloth; PB's results are not realistic enough either, as the logo is distorted and not properly scaled despite of giving parsing input. In contrast, our \modelshort\ can eliminate the distraction from the original cloth without the requirement of any parser information.

\noindent\textbf{Complex Human Posture.}
When facing complex postures, prior solutions cannot generate stable results.
In the third row, where the reference person has one hand on her hip, even though the parsing result seems to be accurate in this instance, the existing PB methods still fail to fit the target clothes naturally to the reference person. PFAFN is not able to generate sharp margin details either. Our \modelshort, on the other hand, can better deal with the complex posture.





\subsection{Quantitative Results}
For virtual try-on, due to the absence of ground truth in the testing scenario, we adopt the Fréchet Inception Distance (FID) \cite{heusel2017gans} as the evaluation metric following \cite{dong2019towards,ge2021parser}, which indicates the similarity between generated images and reference person images. Lower score indicates higher quality of results. 
From the results in Table \ref{tab:fid}, we can tell that our \modelshort\ outperforms both PB and PF methods at different resolution on both datasets. 

\begin{table}[]
\centering
\caption{Quantitative evaluation results of FID. Lower score of FID indicates higher quality of the results.}
\label{tab:fid}
\resizebox{0.8\columnwidth}{!}{
\begin{tabular}{@{}cccc@{}}
\toprule[2pt]
Datasets               & Method      & 256 $\times$ 192        & 512 $\times$ 384         \\ \midrule
\multirow{4}{*}{VITON} & CP-VTON     & 24.45                   & 59.06                    \\
                       & CP-VTON+    & 21.04                   & 51.03                    \\
                       & ACGPN       & 16.73                   & 47.76                    \\
                       & PFAFN       & 10.16                   & 11.57                    \\
                       & \modelshort\ (ours) & \textbf{9.90 (+2.55\%)} & \textbf{9.93 (+14.17\%)} \\ \midrule
\multirow{3}{*}{MPV-Sub}   & WUTON       & 10.89                   & -                        \\
                       & PFAFN       & 10.01                   & -                        \\
                       & \modelshort\ (ours) & \textbf{9.36 (+6.49\%) }                  & -                        \\ \bottomrule[2pt]
\end{tabular}
}
\end{table}

\begin{table}[]
    \centering
    \caption{User study on different datasets. A denotes the baselines, and B denotes our \modelshort.}
    \label{tab:user}
    \resizebox{0.8\columnwidth}{!}{
    \begin{tabular}{@{}cccc@{}}
    \toprule[2pt]
    Dataset                & Baselines   & Res. & Human (A/B)     \\ \midrule
    \multirow{4}{*}{VITON} & CP-VTON  & 256  & \ \ 8.6\%\ / \textbf{91.4\% } \\
                           & CP-VTON+ & 256  & \ \ 6.3\%\ / \textbf{93.7\% } \\
                           & ACGPN    & 256  & 14.9\%\ / \textbf{85.1\%} \\
                           & PFAFN    & 512  & 23.7\%\ / \textbf{76.3\%}\\ \midrule
    \multirow{2}{*}{MPV-Sub}   & WUTON    & 256  & 18.3\%\ / \textbf{81.7\%}\\
                           & PFAFN    & 256  & 24.4\%\ / \textbf{75.6\%}\\ \bottomrule[2pt]
    \end{tabular}
    }
    \end{table}

\subsection{User Study}
To further validate whether the generated clothes look visually real to person, we recruit 20 volunteers in an A / B manner to participate in a user study. The final results of our user study on Table \ref{tab:user} prove that the try-on images generated by \modelshort\ look more realistic to human than other baselines. Detailed settings of user study can be found in Appendix.

\begin{figure}[]
    \centering
    \includegraphics[width=0.95\columnwidth]{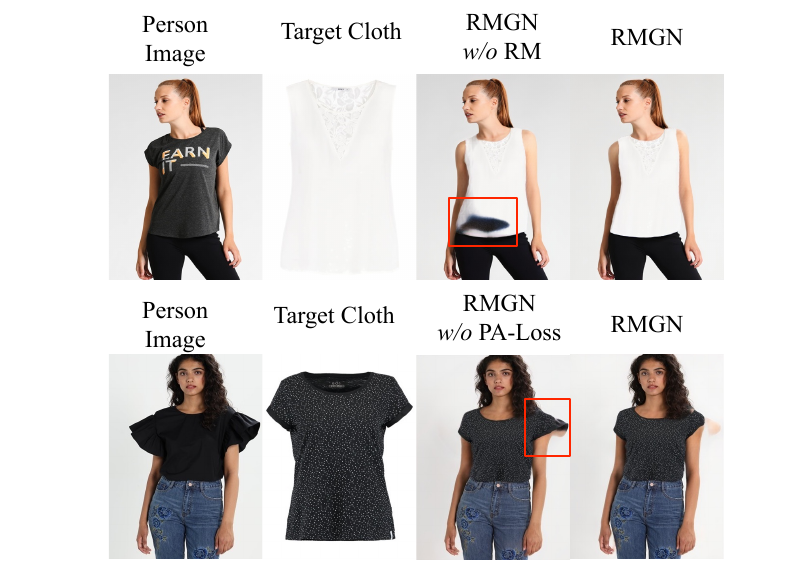}
    \caption{Ablation study on \gate\ (\gateshort) and Posture Awareness Loss (PA-Loss) modules.}
    \label{fig:ablation}
\end{figure}

\begin{figure}[]
  \centering
\begin{minipage}[b]{0.37\columnwidth}
    \centering
    \includegraphics[trim=0 0.7cm 0 0 0,clip,width=\textwidth]{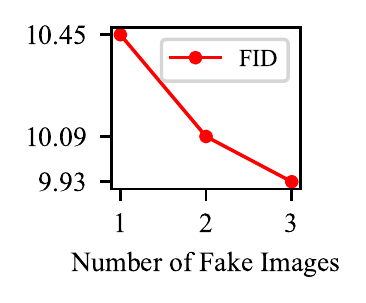}
    \captionof{figure}{The effect of Fake Image Number.}
    \label{fig:fakeimages}
\end{minipage}
\hfill
\begin{minipage}[b]{0.62\columnwidth}
    \centering
\resizebox{\textwidth}{!}{
\begin{tabular}{@{}ccccc@{}}
\toprule[2pt]
Modules                                                          & A            & B            & C            & D            \\ \midrule
Baseline                                                         & $\checkmark$ & $\checkmark$ & $\checkmark$ & $\checkmark$ \\
Multi-level Feature Extractor                                                &              & $\checkmark$ & $\checkmark$ & $\checkmark$ \\
\begin{tabular}[c]{@{}c@{}}Regional Mask \end{tabular}       &              &              & $\checkmark$ & $\checkmark$ \\
\begin{tabular}[c]{@{}c@{}}Posture Awareness Loss\end{tabular} &              &              &              & $\checkmark$ \\ \midrule
FID                                                              & 11.57        & 10.73           & 10.45        & \textbf{9.93}        \\ \bottomrule[2pt]
\end{tabular}
}
\captionof{table}{Ablation Study on VITON (512x384).}
\label{tab:ablation}
  \end{minipage}
\end{figure}


\subsection{Ablation Study}
To evaluate the effectiveness of \textit{\gate\ (\gateshort)} and \textit{posture awareness loss (PA-Loss)}, we further design the ablation study by removing the corresponding modules. The results are shown in the Figure \ref{fig:ablation} where the first row shows the comparison $w/o$ \gateshort\ and the second row shows $w/o$ PA-loss. 
We can observe that when removing the \gate, there is a large black residual in the red rectangle area, which is the remaining of the original cloth. 
By removing the posture awareness loss, \modelshort\ cannot well focus on the posture of the reference person and its performance could be affected by the original cloth.

We also evaluate the effect of different number of fake images sampled for posture awareness loss on \modelshort's performance. Due to the limitation of GPU memories, we only test the number of images as large as 3. As shown in Figure \ref{fig:fakeimages}, when the number of fake images increases, the image quality generated by \modelshort\ also improves. This further shows the effectiveness of the posture awareness loss. 

To further evaluate the effectiveness of each module on the performance improvement, we gradually remove the \textit{posture awareness loss}, \textit{\gate}, and \textit{multi-level feature extractor} from \modelshort. The quantitative results on VITON (512 $\times$ 384) dataset are shown in Table \ref{tab:ablation}. From the results, we can see that all the proposed modules are helpful for the performance improvement. Noticeably, the improvement of the regional mask in FID score is the lowest. This is because the FID scores is a global indicator that cannot imply the defect occurring in a
small area. The regional mask is proposed to focus on the the small defect in high-resolution results, which can be seen form the red rectangle area in Figure \ref{fig:viton}.

\begin{figure}[]
    \centering
    \includegraphics[trim=0 0 0 0,clip, width=0.8\columnwidth]{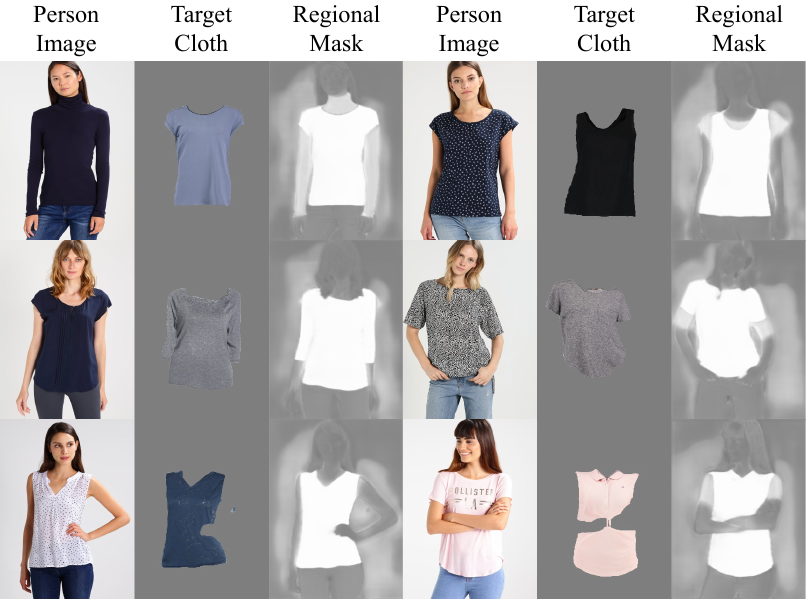}
    \caption{Visualization of \gate.}
    \label{fig:mask}
\end{figure}

\subsection{Case Study}
To deliberate the effectiveness of the proposed \gate\ (\gateshort), we visualize several \gate s extracted from \modelshort\ in Figure \ref{fig:mask}, where the highlighted areas denote the areas of warped cloth and the dark areas denote the areas of reference person. 
In general, \gateshort\ eliminates the interference of incoming features by learning a mask considering the region of the cloth and human body explicitly. It automatically selects the incoming features for synthesized images at the pixel level.
Furthermore, \gateshort\ can adaptively adjust itself to different inputs. From the second row, we can see that when the area of warped clothes is different, \gateshort\ can generate different mask results to synthesize visually natural images. On the left of the second row, when the warped cloth's sleeves are longer, \gateshort\ focuses on the region of the target cloth and preserves it in the synthesized image. On the right of the second row, the warped cloth's sleeves are shorter than the original cloth. \gateshort\ focuses on completely masking the original cloth region and recovering the skin details. Last but not least, from the third row, we can see that \gateshort\ has the ability to identify the human posture, which allows the \modelshort\ to better handle complicated human postures.

\section{Conclusion}
\label{sec:conclusion}

In this paper, we propose a parser-free VTON model called \modelfull\ for generating high-quality pictures in high-resolution, which eliminates the distraction of the reference persons' clothes and can handle complex human postures. In particular, we design a multi-ways generator to separately extract features, and propose a \gate\ to explicitly select incoming features during the feature fusion phase. Furthermore, a posture awareness loss is proposed to focus on the posture information. Extensive experiments on two public datasets show that \modelshort\ outperforms both state-of-the-art PB and PF methods, which indicates that RMGN is potentially favorable in various e-commerce applications due to its photo-realistic results and lightweight deployment. 

\section{Acknowledgements}
This work is supported by National Natural Science Foundation of China (Grant No.62106221).

\bibliographystyle{named}
\bibliography{sections/main.bib}

\end{document}